\documentclass[runningheads]{llncs}

\usepackage[mobile]{eccv}

\usepackage{eccvabbrv}

\usepackage{graphicx}
\usepackage{booktabs}

\usepackage[accsupp]{axessibility}  
\usepackage{algorithm}
\usepackage{algorithmic}

\usepackage[breaklinks,colorlinks,citecolor=eccvblue]{hyperref}

\usepackage{hyperref}
\usepackage{xcolor}

\usepackage{orcidlink}

\begin{document}

\title{Information-Theoretic Constraints for Continual Vision-Language-Action Alignment} 

\titlerunning{Info-VLA}





\author{Libang Zhao\inst{1} \and
Qixin Zeng\inst{2} \and
Hongyin Zhang\inst{1} \and
Donglin Wang\inst{1,3}\orcidlink{0000-0000-0000-0000}} 

\authorrunning{L.~Zhao et al.}

\institute{Westlake University, China \and
University of Southampton, UK \and
\email{wangdonglin@westlake.edu.cn}}

\maketitle

\begin{abstract}
When deployed in open-ended robotic environments, Vision--Language--Action (VLA) models need to continually acquire new skills, yet suffer from severe catastrophic forgetting. We observe that this degradation is related to the deterioration of cross-modal information structure, where dependencies among visual observations, language instructions, and actions progressively diffuse during continual adaptation. But existing continual learning methods fail to preserve such cross-modal information dependencies.
Thus, we propose \textbf{Info-VLA}, an information-preserving continual learning framework that maintains cross-modal information structure through two complementary constraints. \textbf{Replay Anchor Contrastive Learning} constructs stable alignment anchors from a frozen teacher model, preserving cross-modal alignment in the representation space. \textbf{Cross-Modal Mutual Information Maximization} further preserves dependency structure between visual and language representations through mutual information constraints. By jointly preserving historical alignment and cross-modal dependency information, Info-VLA balances stability and plasticity during continual learning. Furthermore, experiments on the LIBERO show that Info-VLA significantly outperforms existing methods in both task retention and adaptation.

 \keywords{Vision-Language-Action model \and Information-Theoretic \and Continual Learning}
\end{abstract}

\section{Introduction}
\label{sec:intro}
Vision-Language-Action (VLA) models have demonstrated remarkable zero-shot generalization ability in robotic manipulation tasks by jointly modeling multimodal perception and behavioral reasoning within a unified neural architecture \cite{kim2025openvla, ma2024survey,Wang2025Unified,Zhao2025CoT-VLA}. 
As robotic applications expand into open-ended environments, models are demanded to support lifelong learning capability, continuously acquiring new skills while retaining previously learned knowledge.
\cite{liu2023libero,Meng2025Preserving,Liu2025Continual}. However, sequential training of VLA models \cite{shi2025continual, wang2024comprehensive} inevitably triggers catastrophic forgetting \cite{french1999catastrophic}, where performance on old tasks degrades sharply as the model adapts to new ones\cite{zhou2025chatvla}. 
More critically, VLA models involve multiple modalities, namely perception, language, and action, making the forgetting problem considerably more complex \cite{Chee2025Multi-Modal,Wang2023A}. In large-parameter models, the alignment of information across different modalities is progressively lost during this process \cite{Zhai2023Investigating,Liu2025Continual}.

\begin{figure}[tb]
\centering
    \includegraphics[height=4.5cm]{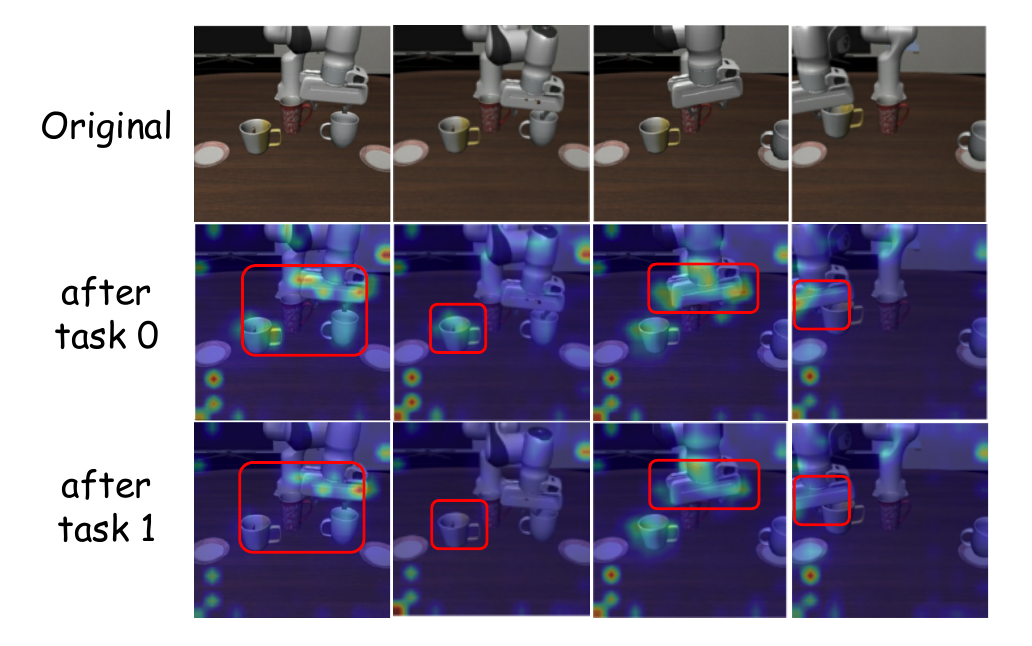}
    \caption{
      Deterioration of cross-modal information structure. After Task 1 ends, compared to the same task, attention diffusion begins.
    }
\label{fig1}
\vskip -0.2in
\end{figure}
Existing mitigation paradigms, including Experience Replay (ER) \cite{rolnick2019experience}, parameter regularization \cite{zhao2024statistical, ahn2019uncertainty}, and architectural expansion \cite{wang2025self,rusu2016progressive,mallya2018packnet}, fail to address this structural decoupling. A common limitation is that they treat VLA learning as task-by-task behavior fitting rather than preserving the historical structure of multimodal information.
Specifically, ER methods focus on approximating historical marginal distributions through raw data recovery, yet they are insufficient for preserving fine-grained cross-modal relationships. 
Regularization-based approaches, such as Elastic Weight Consolidation (EWC) \cite{huszar2018note,kirkpatrick2017overcoming}, limit large weight changes but prove too rigid to accommodate the non-stationary semantic shifts inherent in VLA alignment. Similarly, modular expansion strategies avoid parameter interference at the cost of task isolation, precluding the mixed-modal dependencies essential for generalized behavior. 

In this work, we argue that catastrophic forgetting in VLA models largely arises from the degradation of cross-modal information structure during continual learning. Specifically, as illustrated in Fig.~\ref{fig1}, attention patterns for the same task become increasingly diffused after the model adapts to new tasks, indicating that the conditional dependencies between visual observations and language instructions are progressively distorted.

To address this issue, we propose \textbf{Info-VLA}, an information-theoretic framework that preserves cross-modal information structure while allowing behavioral adaptation.
Info-VLA achieves this goal through two complementary objectives.
\textbf{(1) Replay Anchor Contrastive Learning} introduces stable representation anchors derived from frozen teacher models, preventing newly learned behaviors from overwriting the structural organization of historical representations.
\textbf{(2) Cross-Modal Mutual Information Maximization} explicitly preserves the statistical dependencies between visual and language representations, ensuring that cross-modal relationships remain consistent across task transitions.
Together, these objectives maintain cross-modal information structure during continual learning, enabling VLA models to retain prior knowledge while acquiring new skills.
The primary contributions are as follows:
\begin{itemize}
\item We identify structural distortion of multimodal dependencies as a fundamental cause of catastrophic forgetting in VLA continual learning.
\item We propose Info-VLA, a unified framework that preserves cross-modal structural stability through replay-anchored contrastive learning and mutual information–based distillation. 
\item Extensive evaluations show that Info-VLA establishes a new state-of-the-art in both task retention and adaptation efficiency.
\end{itemize}

\section{Related Work}
\label{author info}
\subsection{Vision-Language-Action Models}
Vision--Language--Action learning has emerged as a prominent and rapidly advancing direction in robot agents for open-ended world interaction~\cite{ma2024survey,sapkota2025vla,din2025vla}. Transformer-based approaches including ACT~\cite{zhao2023learning}, RT-1~\cite{brohan2022rt}, RT-2~\cite{zitkovich2023rt}, and HPT~\cite{wang2024scaling} leverage sequence modeling to represent state--action--reward trajectories for improved decision-making. Video generation with inverse kinematics methods such as UniPi~\cite{du2023learning} and RoboDreamer~\cite{zhou2024robodreamer} adopt a two-stage pipeline of motion video generation followed by action recovery. LLM--diffusion hybrid methods such as Octo~\cite{team2024octo} combine multimodal language representations with diffusion-based action generation. As for MoE-based frameworks, the $\pi_0$ family~\cite{black2024pi0} reformulates action generation as Flow Matching, while $\pi_{0.5}$~\cite{black2025pi05} further achieves open-world generalization through heterogeneous task co-training. Unlike prior approaches that focus on single-phase training or task-specific generalization, 
our method addresses catastrophic forgetting and cross-modal misalignment in VLA continual 
learning via information-theoretic constraints.

\subsection{Continual Learning in VLA Model}
While Continual Imitation Learning mitigates catastrophic forgetting via parameter expansion or task decomposition, these approaches struggle to scale to large VLA models due to either parameter explosion or computational overhead ~\cite{zhao2024statistical,huszar2018note,malagon2025self,rolnick2019experience,zhao2024statistical,huszar2018note,malagon2025self}. Recent work proposes Stellar VLA~\cite{wu2025continually}, a knowledge-driven continual learning framework that enables VLA models to sequentially learn new manipulation tasks through self-supervised knowledge evolution and knowledge-guided expert routing.
Recent work~\cite{li2025metavla} put forward a unified post-training framework that enables continual adaptation of VLA models to diverse tasks through context-aware meta co-training, reducing task-specific fine-tuning costs while improving generalization. CRL-VLA~\cite{zeng2026crlvlacontinualvisionlanguageactionlearning} addresses 
the stability-plasticity dilemma in continual robotic learning via a dual-critic 
architecture with goal-conditioned value formulation.
Unlike these methodologies, our work focuses on cross-modal alignment preservation in VLA continual learning, ensuring that the conditional dependencies between vision, language, and action modalities are maintained across sequential task adaptation.
\subsection{Contrastive learning in VLA Model}
 Contrastive learning enables large multimodal model like VLA to learn robust and generalizable representations by capturing the intrinsic alignment between visual, linguistic, and action modalities~\cite{chen2020simple,Wang2023Exploring,Chen2024Vision-Language,Chen2024Vision-Language}.
Specifically, addressing the limited sensitivity of VLA models to robotic signals recent work proposes that enhances control-relevant representation learning by aligning embeddings with proprioceptive states using relative state distances as soft supervision~\cite{Syed2025ExpReS-VLA:}.
VLA-R~\cite{seong2025vla} aligns vision-language embeddings with action embeddings through vision-action contrastive learning, enabling effective open-world reasoning and action retrieval. Different from these works, we focus on leveraging contrastive learning to explicitly maintain cross-modal alignment across sequential tasks, preventing the degradation of VLA dependencies during continual learning.

\section{Preliminaries}
We consider a VLA policy that maps multimodal observations and language instructions to a sequence of robot actions. At time step $t$, the observation is defined as
$o_t = [I^0_t, \dots, I^n_t, q_t]$,
where $I^i_t$ denotes the image captured by camera $i$, and $q_t$ represents the robot proprioceptive state, including joint angles, gripper pose, torso height, and base velocity. The high-level task instruction is denoted as $l$. The policy models a joint conditional distribution over a sequence of action chunks $a_{t:t+H}$ and an intermediate subtask description $\hat{l}$,
$\pi_\theta(a_{t:t+H}, \hat{l} \mid o_t, l).$
Here $a_{t:t+H}$ denotes a horizon-$H$ action sequence starting from time $t$, and $\hat{l}$ is a discrete textual variable representing the inferred subtask. The joint distribution can be factorized as:
\[
\pi_\theta(a_{t:t+H}, \hat{l} \mid o_t, l)
=
\pi_\theta(a_{t:t+H} \mid o_t, \hat{l})
\pi_\theta(\hat{l} \mid o_t, l).
\]

The first term models low-level action execution conditioned on the observation and inferred subtask, while the second term captures high-level subtask inference conditioned on the observation and the overall instruction.

Moreover, in the continual learning setting, we consider a sequence of tasks $\mathcal{T}_0, \mathcal{T}_1, \dots, \mathcal{T}_n$,
where the policy must adapt to new tasks while preserving previously acquired capabilities. For each task $\mathcal{T}_k$, demonstrations are provided as $D_k = \{(o_t^k, a_t^k, l_k)\}_{t=1}^{T_k}$. At the base stage ($k=0$), only dataset $D_0$ is available. During training process ($k>0$), training has access only to the current dataset $D_k$ and a memory buffer $\mathcal{M}$ storing limited past samples, while the remaining historical data are unavailable.

\begin{figure*}
\centering
  \includegraphics[height=4.4cm]{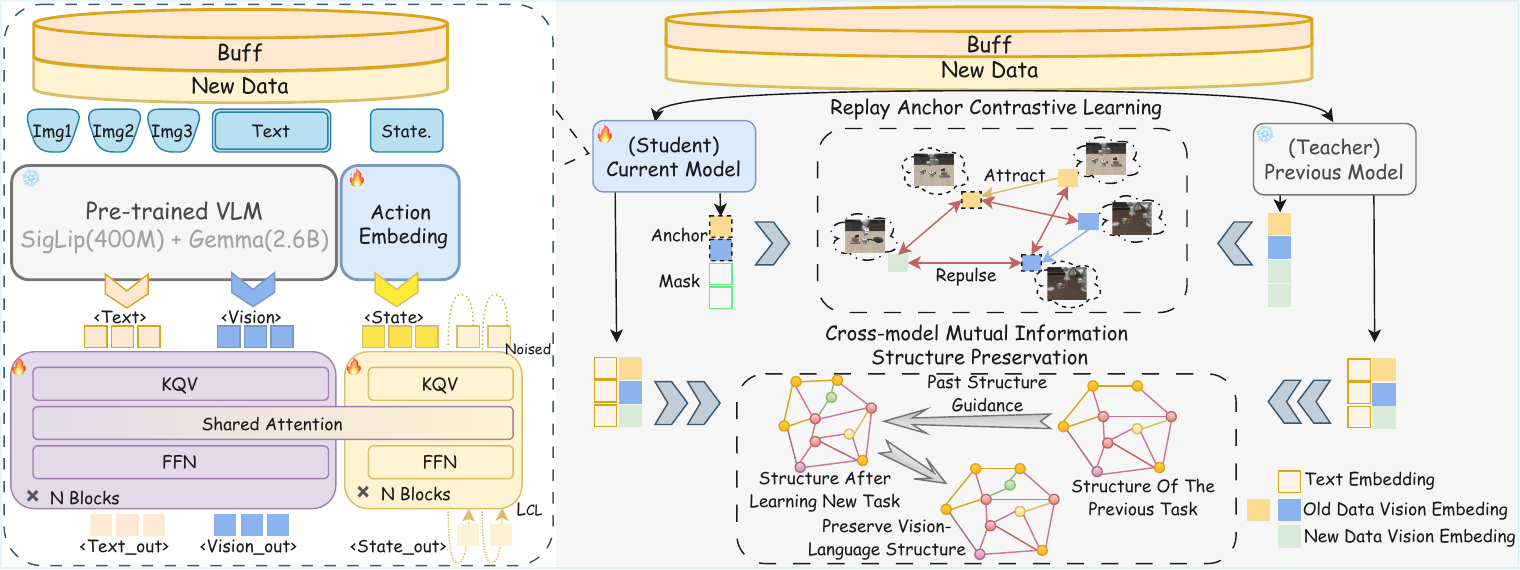}
    \caption{
      The proposed architecture consists of two main components. Left: this part depicts the model’s data flow, covering input feature composition, prediction generation, and the structured action space representation. Right: this section illustrates key approach, leveraging replay-anchored contrastive learning to retain historical task knowledge and vision–language mutual information regularization to enforce cross-modal structural consistency.
    }
    \label{main}
\vskip -0.1in
\label{fig:mi-collapse}
\end{figure*}

\section{Methodology}
In this section, we first introduce the core principles of Info-VLA for preserving cross-modal alignment. We then present its two key components, followed by the overall training objective and pipeline.
\subsection{Structural Principles of Info-VLA}
Empirically, as shown in Fig.~\ref{fig1}, the first row depicts the original observations of the task, and the second row shows the corresponding cross-modal attention maps after Task 0 training.
At this stage, high-response regions are mainly concentrated on instruction-relevant objects and interaction areas, such as the target cups, containers, and the end-effector vicinity, with the highlighted regions exhibiting relatively compact and consistent attention patterns. 
 In contrast, the third row shows the attention maps after Task 1 training, revealing a clear diffusion effect: high-response regions spread from task-relevant targets to irrelevant background areas, while responses on key objects weaken and fragment.
 Continual adaptation introduces interference: the conditional dependencies between visual and linguistic modalities diffuse across modality boundaries and are not stably preserved. Consequently, task-discriminative visual cues are increasingly corrupted, reducing the model's ability to extract and localize instruction-relevant information. For example, when the instruction refers to grasping a cup, attention may partially drift toward background textures, while for instructions emphasizing object placement, attention may shift from the target container toward the robot arm body or the tabletop.
 
This observation indicates that the significant challenge in continual VLA learning is the lack of a stable source of alignment together with the progressive corruption of cross-modal dependency structure. 
Therefore, an effective continual cross-modal alignment requires two essential conditions: stable reference anchors that preserve historically established correspondences, and explicit structural constraints that maintain these correspondences during subsequent updates. 
We present Info-VLA, a replay-based distillation framework for continual learning in VLA models, as shown in Fig.~\ref{fig:mi-collapse}. After completing training on each task, the model is frozen and used as the \textbf{teacher} model for the subsequent task stage. During new task training, the current \textbf{student} model preserves previously established cross-modal structures through two key components: Replay-Anchored Contrastive (RAC) learning and Cross-Modal Mutual Information (CMI) preservation.

\subsection{Replay Anchor Contrastive }
Before enforcing cross-modal alignment constraints, representational interference between new and historical tasks must first be suppressed to achieve effective separation of task-specific representations. Such separation reduces overlap between task feature spaces and provides a more stable visual latent space for subsequent cross-modal alignment, establishing a structurally consistent foundation for further optimization.


Based on this analysis, the RAC learning framework is introduced to sustain model plasticity. Unlike traditional contrastive learning methods that rely on data augmentation or multi-view sampling, RAC derives anchor representations from a frozen teacher model, providing temporally consistent and stable reference benchmarks. As illustrated on the right side of Fig.~\ref{fig:mi-collapse}, RAC integrates contrastive mechanisms with a replay anchor strategy, utilizing historically stored samples as references to prevent the overwriting of previously learned representations during new task training.
Crucially, the anchor representations from the frozen teacher model inherently preserve the visual–linguistic alignment structure established during pre-training. This allows the contrastive objective to impose task-level discriminative constraints while providing a stable supervisory source for cross-modal feature alignment.

As a result, newly acquired behaviors do not disrupt reasoning grounded in historical representation coordinates. Specifically, at the end of each training stage, a teacher model $\pi_{\theta^-}$ is created by copying the student model. Its parameters remain frozen during subsequent training. Replay buffer samples are processed by the teacher model $f_{\theta-}(\cdot)$ to extract latent visual embeddings $V_{Anchor}^{q_t^i,l^i}= f_{\theta-}(o_t^i,l^i)$, which serve as anchors. These embeddings are obtained from the cross-attention layer that fuses visual, task description $l$, and proprioceptive state $q_t$.
Positive samples $V_{pos}^{o_t^i,l^i} = f_{\theta}(o_t^i,l^i)$ are generated by feeding the same replay samples into the student model $f_{\theta}(\cdot)$, ensuring that positive pairs correspond to representations of identical trajectories under both the teacher and student models. Negative samples $V_{neg}^{o_t^j,l^j} = f_{\theta}(o_t^j,l^j) (j \neq i)$ are defined as all other student representations within the batch, including samples drawn from both previous and current tasks. This construction serves two key purposes. First, positive pairs strictly correspond to identical trajectories, thereby avoiding semantic ambiguity. Second, samples from new tasks naturally act as negative examples, promoting separation between old and new task representations in the latent space and reducing representational interference.
Based on this design, an asymmetric contrastive learning objective is adopted:
{
\begin{equation}
\mathcal{L}_{\mathrm{RAC}}=E_{(o,l)\sim {D_t\cup \mathcal{M}}} \left[-\log \frac{\exp \left(\left\langle V^{o_t^i,l^i}, V_{Anchor}^{o_t^i,l^i}\right\rangle / \tau\right)}{\sum_{j \in \mathcal{B}} \exp \left(\left\langle V^{o_t^i,l^i}, V_{Anchor}^{o_t^j,l^j}\right\rangle / \tau\right)}\right] .
\end{equation}
}
where $\tau$ denotes the temperature parameter.
Given a subset 
$\mathcal{B}$ of old-task samples within a mini-batch, the contrastive loss is defined as follows. This objective encourages the student model’s visual representations $V_{pos}^{o_t^i,l^i}$ for old-task samples to align with their corresponding teacher anchors $V_{Anchor}^{o_t^i,l^i}$, while simultaneously increasing their separation from other representations in the batch, particularly those associated with new-task samples.  By guiding new task representations to converge toward the alignment manifold defined by the anchors, RAC alleviates inter-task interference and establishes a structurally consistent feature foundation for subsequent mutual information-based cross-modal distillation.

\subsection{Cross-Modal Mutual Information}
Contrastive learning constrains pairwise distances among latent representations, stabilizing the optimization trajectory during continual adaptation. However, such pairwise alignment mechanisms do not explicitly preserve global dependency structures across modalities between model states at different time steps. In continual cross-modal learning, catastrophic forgetting manifests not only as representational drift but also as systematic distortion of the joint action–language distribution. Therefore, cross-modal structural consistency between the prior teacher model and the current student model must be explicitly maintained.

To address this limitation, cross-modal structure is preserved by maintaining the statistical dependency between visual and textual representations under language conditioning, rather than enforcing pointwise activation imitation. The teacher network produces visual–text latent representations ${Z}^{{old}}$, while the student network generates corresponding representations ${Z}^{{new}}$. The formulation is illustrated in the CMI section on the right side of the Fig.~\ref{fig:mi-collapse}.
The teacher representation space encodes rich semantic dependency structures, including inter-sample similarities, class decision boundaries, and hierarchical conceptual relationships. Such structural information is not reflected in any individual activation value. Instead, it is distributed across the statistical properties of the overall representation distribution. By maximizing the mutual information) between the teacher’s visual–language latent representations $Z^{old}$ and the student’s corresponding representations $Z^{new}$, structural guidance is imposed on the student’s representation space. This objective preserves cross-modal structural consistency throughout the continual learning process:
\vskip -0.05in
{
\small
\begin{align}
\label{eq2}
\mathcal{L}_{MI} = -I\left(Z^{new} ; Z^{old}\right) = -D_{KL}\left( p(Z^{new} ,Z^{old})\|p(Z^{new} )p(Z^{old})\right),
\end{align}
}
where $D_{KL}$ denotes the Kullback–Leibler (KL) divergence. The mutual information is zero if and only if $Z^{old}$ and $Z^{new}$ are statistically independent. Conversely, a large mutual information value indicates that the joint distribution $p(Z^{old},Z^{new})$ deviates substantially from the product of their marginal distributions. In this case, sample pairs regarded as semantically similar by the teacher tend to exhibit similar statistical behavior in the student’s representation space, and the teacher’s category-boundary structures are implicitly transferred to the student. The joint distribution $p(Z^{old},Z^{new})$ is parameterized by concatenating the two representations and passing the result through a two-layer Multi-Layer Perceptron (MLP): $p(Z^{old},Z^{new}) = MLP(F([z^{old},z^{new}]))$. The marginal distributions are estimated independently as:

\begin{align}
\label{eq2}
\nonumber
p(Z^{old}) = F(z^{old}), \\ p(Z^{new}) = F(z^{new}),
\end{align}
where $F(\cdot)$ denotes a shared projection function, $F(z) = softmax(V\cdot L^\top)$, $V$ and $L$  denote the visual and language projection matrices. This design makes the mutual information estimator fully end-to-end differentiable and seamlessly integrates it into the unified training framework.

Since mutual information maximization does not impose explicit constraints on the marginal distributions, it may lead to degenerate solutions in which the marginals become overly sharp without fully collapsing. To mitigate this issue, a marginal consistency regularization term is introduced to align the global behavioral-intention distributions of the student and teacher models. This regularization enforces consistency in their global decision tendencies, thereby preserving the coherence of the student’s overall reasoning process. The corresponding regularization term is defined as follows:
\begin{align}
\label{eq2}
\mathcal{L}_{MC} = D_{KL}(p(Z^{new}) \| p(Z^{old})).
\end{align}

This marginal consistency term encourages the student model to maintain similar global distributional statistics as the teacher model while allowing flexibility at the instance level. Importantly, this regularization operates at the distribution level rather than enforcing pointwise alignment, thereby preserving semantic structure without restricting model plasticity. The unified structural preservation objective is defined as follows:
\begin{align}
\label{eq2}
\mathcal{L}_{CMI} = \mathcal{L}_{MI} + \mathcal{L}_{MC}.
\end{align}

\subsection{Overall Objective and Training Pipeline}
\begin{algorithm}[tb]
\caption{Training Procedure of Info-VLA}   
\begin{algorithmic}[1]  
\label{algo}
\REQUIRE 	 Pretrained VLA policy $\pi_{\theta}$, tasks $\{\mathcal{T}_k\}_{k=0}^K$.
\STATE \textbf{for} $k = 0$ to $K$:
    \STATE \hspace{1em} \textbf{if } $k = 0$:
        \STATE \hspace{2em} Train $\pi_{\theta_0}$ on $\mathcal{T}_0$  by Eq.~\ref{eqCL}
        \STATE \hspace{2em} Freeze $\pi_{\theta_0}$ 
         \STATE \hspace{2em} Each task stores a single trajectory to $\mathcal{M}_{\mathrm{old}}$
    \STATE \hspace{1em} \textbf{else}: 
        \STATE \hspace{2em} $\mathcal{D}_k = \mathcal{T}_k \cup \mathcal{M}_{old}$
        \STATE  \hspace{2em} \textbf{for} each training iteration:
            \STATE \hspace{3em} Sample batch $\mathcal{B} \sim \mathcal{D}_k$,
            \STATE \hspace{3em} Compute losses $\mathcal{L}$ by Eq.~\ref{eq7}
            \STATE \hspace{3em} Update $\theta$ via gradient descent on $\mathcal{L}$
    \STATE \hspace{2em} Freeze $\pi_{\theta_k}$ and Store one trajectory to $\mathcal{M}_{\mathrm{old}}$
\STATE Continually adapted VLA policies $\{\pi_{\theta_k}\}_{k=0}^K$
\end{algorithmic}
\end{algorithm}

We performs continual post-training on the policy $\pi_{0.5}$ that is pre-trained on large-scale web data. To maintain the stability of the multimodal backbone during continual adaptation, only the parameters of the action head are updated while the visual--language encoder remains frozen.
The base training objective follows the flow-matching formulation used in the original policy learning process:
\begin{align}
\label{eqCL}
\mathcal{L}_{CL} =
\left\|
\omega - \mathbf{a}_{t:t+H} -
f_{\theta}^{a}
\left(
\mathbf{a}_{t:t+H}^{\tau,\omega},
\mathbf{o}_{t},
\ell
\right)
\right\|^{2},
\end{align}
where $\tau \in [0,1]$ denotes the flow-matching time index, $\omega \sim \mathcal{N}(0,I)$ represents Gaussian noise, and $(a,o,\ell) \in D \cup \mathcal{M}$ corresponds to samples drawn from the current dataset and the replay memory. The prediction $y_{1:H}^{a}=f_{\theta}^{a}(\mathbf{a}_{t:t+H}^{\tau,\omega},\mathbf{o}_{t},\ell)$ is produced by a lightweight action expert.
Therefore, the overall training objective integrates the Replay Anchor Contrastive loss and the Cross-Modal Mutual Information loss as:
\begin{align}
\label{eq7}
\mathcal{L} =
\mathcal{L}_{CL}
+
\lambda_1 \mathcal{L}_{\mathrm{RAC}}
+
\lambda_2 \mathcal{L}_{\mathrm{CMI}},
\end{align}
where $\lambda_1$ and $\lambda_2$ control the relative contributions of the contrastive distillation loss $\mathcal{L}_{\mathrm{RAC}}$ and the cross-modal mutual information loss $\mathcal{L}_{\mathrm{CMI}}$.

The complete continual learning procedure of Info-VLA is summarized in Algorithm~\ref{algo}. The policy is first trained on the initial task $\mathcal{T}_0$ using Eq.~\ref{eqCL}, after which the resulting policy is frozen as the teacher model. For subsequent tasks, training batches are sampled from the union of the current task data and the replay memory, and the policy parameters are updated by optimizing the full objective in Eq.~\ref{eq7}.

\section{Evaluation}
Having presented Info-VLA for mitigating catastrophic forgetting and preserving cross-modal consistency, we now evaluate its effectiveness through a comprehensive experimental analysis, including detailed setup, performance comparisons, and ablation studies.
\subsection{Experimental Setup}
\paragraph{Implementation Details.}
Experiments employ the LIBERO benchmark~\cite{liu2023libero}, specifically designed for continual learning. The model employs $\pi_{0.5}$~\cite{black2025pi05} as the pre-trained foundation model. Several configurations are defined using the notation $Bi\text{-}kNm$. This notation indicates that the policy is first trained on $i$ base tasks, followed by the incremental addition of $m$ tasks at each step for a total of $k$ steps. Evaluation is conducted on two benchmarks: (1) 10 sequential long-horizon tasks from LIBERO-Long, and (2) the first five tasks from LIBERO-Goal. Both benchmarks require the robot to understand natural language instructions and execute multi-step actions, including object picking, drawer manipulation, and knob rotation. $B5\text{-}5N1$ is adopted for LIBERO-Long, and $B0\text{-}5N1$ is adopted for LIBERO-Goal.  For past tasks, one trajectory is randomly stored for each task.
The hyperparameter $\lambda_1$ and $\lambda_2$ are set to 0.1. Base training with Base task runs for 3,000 iterations in total. Each incremental step is trained for 600 iterations.

\paragraph{Metrics.}
Performance is evaluated using five quantitative metrics: ~\cite{liu2023libero,diaz2018forgetting}: the area under the accurate-rate curve(AUC): $AUC = \frac{1}{N} \sum_{i=1}^N \left( \frac{1}{N-i+1} \sum_{j=i}^N R_{i,j} \right)$, the forward transfer (FWT):  $FWT = \frac{1}{N-i+1} \sum_{i=1}^N R_{i,i}$,  negative backward transfer(BNT): $NBT = \frac{1}{N-1} \sum_{i=1}^{N-1} \left( \frac{1}{N-i} \sum_{j=i+1}^N R_{i,i} - R_{i,j} \right)$,  final average accurate(FAA) rate $FAA = \frac{1}{N} \sum_{i=1}^N R_{i,N} $ and the average accurate(AA): $AA = \frac{1}{N} \sum_{j=1}^N \left(\frac{1}{j} \sum_{i=1}^j R_{i,j} \right)$. Suppose that after learning $j$ tasks, the success rate on the 
i-th task is denoted as $R_{i,j}$. Overall, AUC reflects overall performance, FWT measures learning ability, NBT quantifies forgetting, FAA denotes all task final mean accuracy, and AA denotes mean task accuracy. 
\begin{table*}[t]
  \caption{Performance comparison of different methods on the LIBERO-Long $B5\text{-}5N1$ benchmark. Old denotes the average success rate on previously learned tasks, All denotes the average success rate over all encountered tasks, and AA summarizes average success across the entire continual learning sequence. Bold denotes best performance.}
  \label{table1}
  \begin{center}
        \begin{tabular}{l|cccccccccccc}
          \toprule
          methods& Base & \multicolumn{2}{c}{Task1}& \multicolumn{2}{c}{Task2}& \multicolumn{2}{c}{Task3}& \multicolumn{2}{c}{Task4}& \multicolumn{2}{c}{Task5} & AA \\
           & & Old & All& Old & All& Old & All& Old & All& Old & All &\\
          \midrule
          multitask  &  &\multicolumn{2}{c}{92.4}&\multicolumn{2}{c}{92.4}&\multicolumn{2}{c}{92.4}&\multicolumn{2}{c}{92.4}&\multicolumn{2}{c}{92.4}    \\ \hline
          Sequential & 80.8& 8.8& 22.3& 24.0& 32.0& 0.0& 11.5& 9.0& 18.4& 0.8 & 10.0 & 29.2\\
          EWC & 80.8& 7.2& 10.3& 8.0& 12.0& 0.0& 6.0& 9.7& 19.3& 6.2 & 15.4 & 24.0\\
          ER  & 80.8& 63.6& 64.3& 75.6& 78.0& 68.3& 72.3& 67.0& 69.8& 64.0 & 66.6 & 72.0\\
          Ours & 80.8& \textbf{72.8}& \textbf{75.0}& \textbf{83.3}& \textbf{82.3}& \textbf{71.4}& \textbf{74.5}& \textbf{84.7}& \textbf{86.0}& \textbf{71.0} & \textbf{73.6}& \textbf{78.7}\\
          \bottomrule
        \end{tabular}
  \end{center}
  \vskip -0.2in
\end{table*}

\begin{table*}[t]
  \caption{Performance comparison of different methods on the LIBERO-Goal $B0\text{-}5N1$ benchmark. Continual learning proceeds directly from the original pre-trained model. Old denotes the average success rate on previously learned tasks, All denotes the average success rate over all encountered tasks, and AA summarizes average success across the entire continual learning sequence. Bold denotes best performance.}
  \label{tablegoal}
  \begin{center}
        \begin{tabular}{l|cccccccccc}
          \toprule
          methods& \multicolumn{1}{c}{Task1}& \multicolumn{2}{c}{Task2}& \multicolumn{2}{c}{Task3}& \multicolumn{2}{c}{Task4}& \multicolumn{2}{c}{Task5} & AA \\
           &   All& Old & All& Old & All& Old & All& Old & All &\\
          \midrule
          multitask   &\multicolumn{1}{c}{98.4}&\multicolumn{2}{c}{98.4}&\multicolumn{2}{c}{98.4}&\multicolumn{2}{c}{98.4}&\multicolumn{2}{c}{98.4}    \\ \hline
          Sequential& 94.0 & 2.0& 40.0& 6.0& 34.0& 0.0& 22.5& 1.0 & 20.0 & 42.1\\
          EWC & 96.0 & 12.0& 49.0& 9.0& 24.0& 3.0& 25.0& 4.0 & 21.0 & 43.0\\
          ER  & 96.0 & 48.0& 63.0& 52.0& 63.3& 44.7& 56.0& 43.0 & 53.6 & 67.2\\
          Ours & \textbf{98.0}&   \textbf{68.0}& \textbf{73.0}& \textbf{56.0}& \textbf{67.3}& \textbf{54.7}& \textbf{64.0}& \textbf{55.0} & \textbf{64.0}& \textbf{73.3}\\
          \bottomrule
        \end{tabular}
  \end{center}
  \vskip -0.1in
\end{table*}

\paragraph{Baselines.}
\textbf{Multitask}~\cite{liu2023libero} jointly trains on all tasks with full access to historical data, serving as the upper bound for continual learning.
\textbf{Sequential}~\cite{liu2023libero} fine-tunes all parameters per task without any forgetting mitigation, typically suffering severe catastrophic forgetting.
\textbf{ER}~\cite{rolnick2019experience} uses one stored trajectory per task as replay data, mixed with new task data for continual training.
\textbf{EWC}~\cite{huszar2018note} mitigates forgetting via Fisher Information Matrix-based quadratic penalties on important parameters.

\begin{figure}[tb]
  \centering
  
  \includegraphics[height=5.8cm]{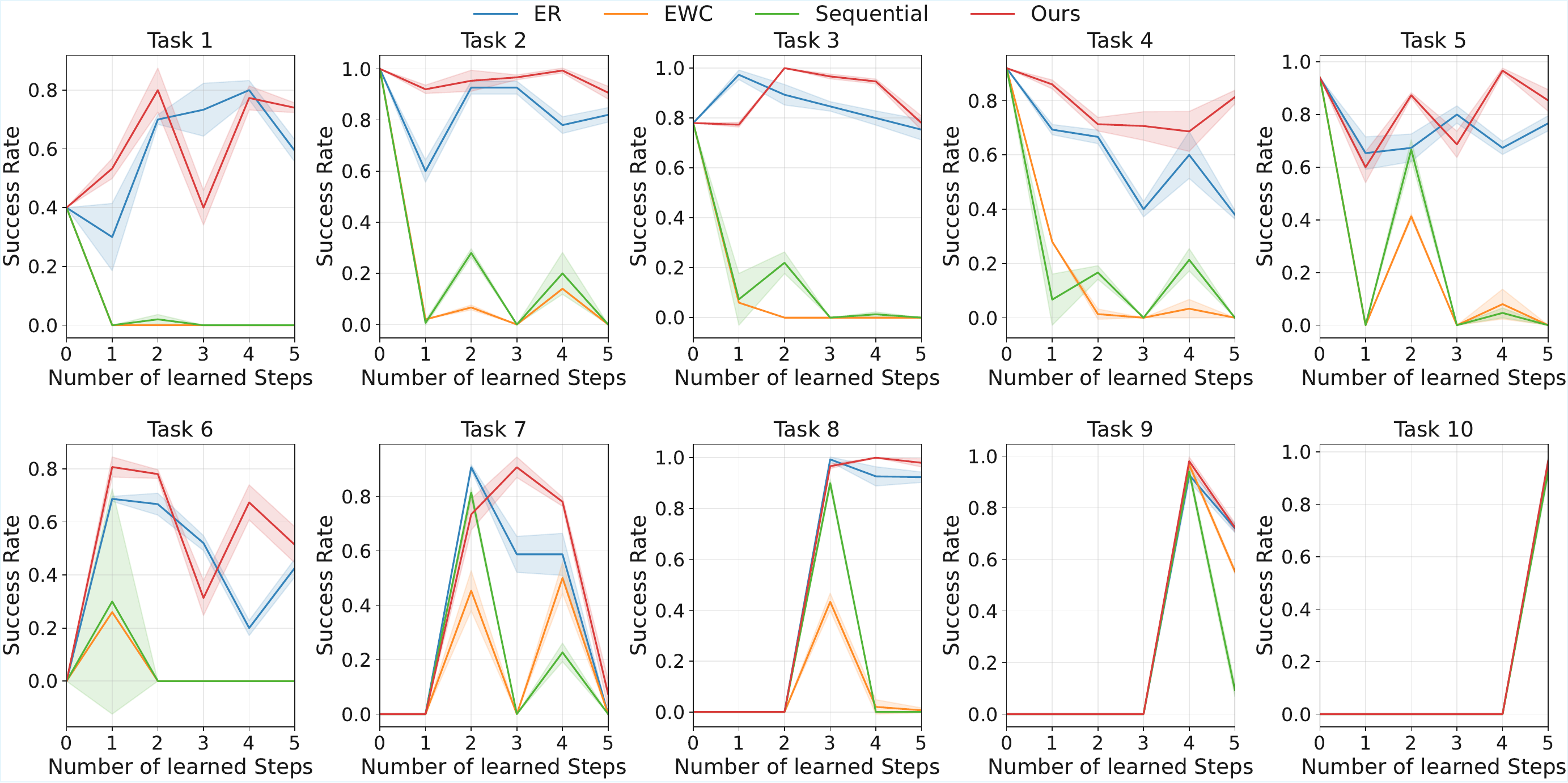}
  \caption{The evolution of success rates for our method compared with five baseline methods on the LIBERO-Long $B5\text{-}5N1$ benchmark. Solid curves represent the average success rates across three runs with different random seeds, and the shaded areas correspond to standard deviation. 
  }
  \vskip -0.1in
 \label{fig3}
\end{figure}

\begin{figure*}[t]
\centering
    \includegraphics[height=4cm]{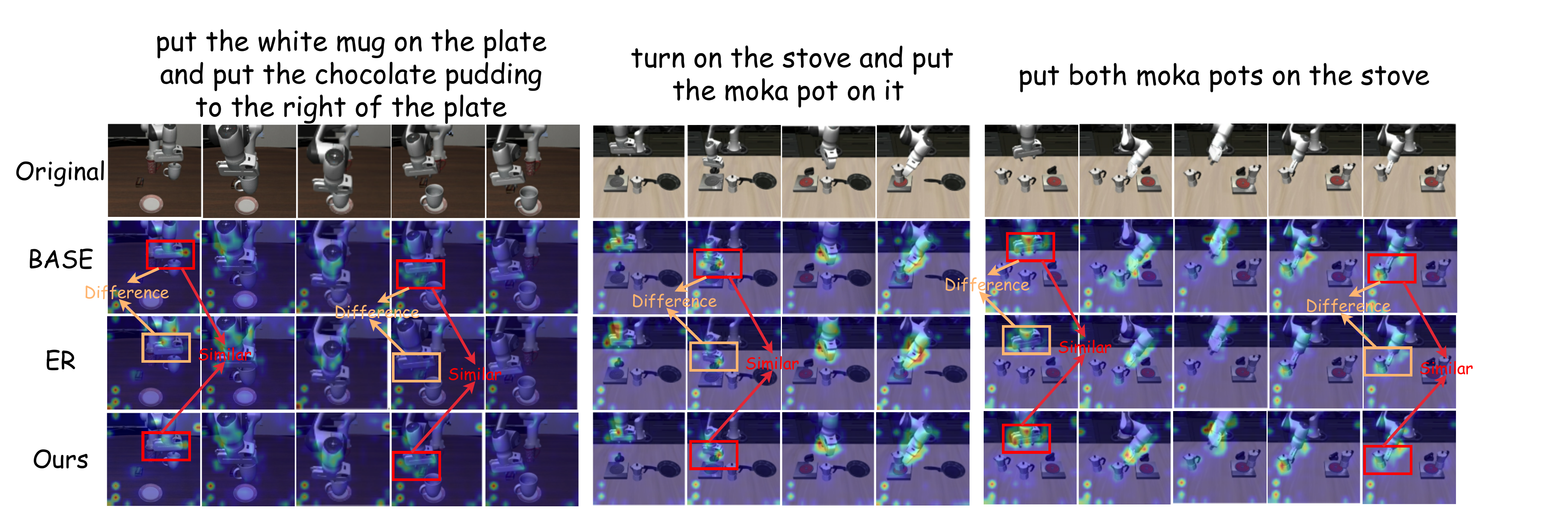}
    \caption{
      Ablation Study on Mutual Information Structure Preservation.
BASE denotes the representation structure immediately after learning the target task. ER and Ours illustrate the representation structure of the same task after subsequent training on a new task. Red boxes indicate similarity, while orange boxes indicate dissimilarity.
    }
\label{fig4}
\end{figure*}

\begin{figure}[tb]
\centering
    \includegraphics[height=3cm]{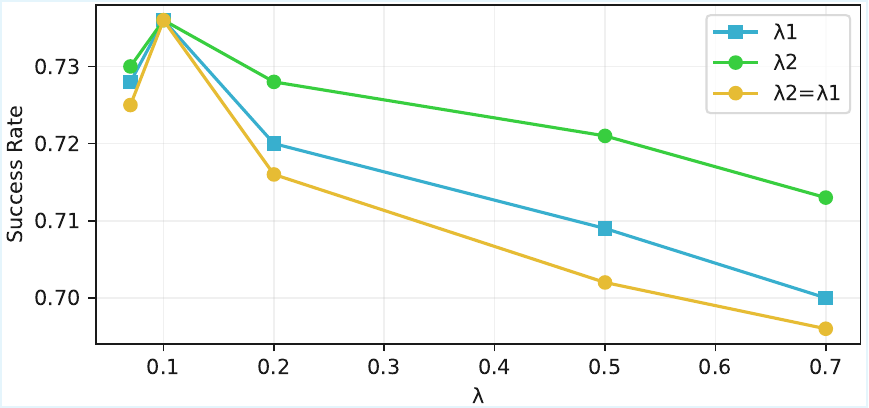}
    \caption{
      Ablation study.  $\lambda_1$ and $\lambda_2$ are hyperparameters that balance the contributions of the RAC and CMI losses. When $\lambda_1$  is varied, $\lambda_2$ is fixed at 0.1 , $\lambda_2$  is varied, $\lambda_1$ is fixed at 0.1. Final average accuracy with varying values of $\lambda$.
    }
    \vskip -0.2in
\label{fig5}
\end{figure}

\subsection{Main Results}
Firstly, to evaluate the overall performance of Info-VLA, we conduct experiments on the Libero-Long $B5\text{-}5N1$  benchmark, which contains ten sequential tasks. The first five tasks serve as base tasks for initial training, followed by five incremental tasks introduced one at a time. Table~\ref{table1} compares our method with baseline approaches, showing that Info-VLA consistently achieves the highest overall performance across all stages. Compared to the best-performing ER baseline, Info-VLA improves old-task, overall, and phase-wise accuracy by approximately 6–9\%, demonstrating a superior stability–plasticity trade-off. Furthermore, the effectiveness of the proposed method is evaluated under a setting where continual learning is performed without task-specific adaptation. Results on the LIBERO-Goal $B0\text{-}5N1$  benchmark are reported in Table~\ref{tablegoal}. Compared to the best-performing ER baseline, Info-VLA improves old-task, overall, and phase-wise accuracy by approximately 5–10\%.

\begin{table}[t]
  \caption{Ablation study of individual components. Evaluated on the LIBERO-Long $B5\text{-}5N1$ benchmark. Ablation experiments are conducted to analyze the contributions of Replay Anchor Contrastive (RAC) Learning and Cross-Modal Mutual Information (CMI) Maximization. The best success rate is marked in red.}
  \label{table2}
  \vskip -0.1in
  \begin{center}
      \begin{sc}
        \begin{tabular}{ll|cccc}
          \toprule
          RAC & CMI & AUC ↑ & FWT ↑ & NBT ↓ & FAA ↑ \\
          \midrule
          $\times$ & $\times$ & 73.9 & 88.4 & \textcolor{red}{-4.3} & 66.6 \\
          $\surd$  & $\times$ & 78.5 & 89.4 & -0.4                  & 73.5 \\
          $\times$ & $\surd$  & 79.3 & 88.8 & -4.1                  & 68.6 \\
          $\surd$  & $\surd$  & \textcolor{red}{81.0} & \textcolor{red}{91.2} & -1.0 & \textcolor{red}{73.6} \\
          \bottomrule
        \end{tabular}
      \end{sc}
  \end{center}
  \vskip -0.2in
\end{table}

Secondly, to analyze learning dynamics during incremental stages, we examine per-task success rates across all steps (Fig.~\ref{fig3}). The results indicate that Info-VLA substantially mitigates catastrophic forgetting and outperforms existing baselines on most tasks. Performance on some tasks improves in later stages, as indicated by negative NBT values (Table~\ref{table2}), likely due to shared trajectories between early and later tasks.

Finally, to assess how well Info-VLA preserves cross-modal representations, we compare vision–language–action cross-attention patterns under three settings: after base-task training, after the first incremental step using ER, and after the first incremental step using Info-VLA (Fig.~\ref{fig4}). Visualization results demonstrate that Info-VLA better maintains the original attention patterns from the base stage compared to ER, confirming the effectiveness of integrating contrastive learning with mutual information optimization for preserving cross-modal structure.

\subsection{Ablation Study}
We conduct ablation studies on both hyperparameters and model components.
\textbf{(1) Hyperparameter analysis.} Results show FAA on the LIBERO-Long $B5\text{-}5N1$ benchmark.
We analyze the influence of the weighting coefficients $\lambda_1$ and $\lambda_2$. As shown in Fig.~\ref{fig5}, we first set $\lambda_1=\lambda_2$ to determine a reasonable operating point. The best performance is obtained when $\lambda_1=\lambda_2=0.1$. We then vary $\lambda_2$ while fixing $\lambda_1=0.1$, and vice versa. In both cases, the optimal performance remains at $0.1$, indicating that the method is relatively stable with respect to these hyperparameters. Therefore, $\lambda_1=\lambda_2=0.1$ is used in all experiments.
\textbf{(2) Component ablation.}
We use ER as the baseline and evaluate the two key components, RAC and CMI, individually. As shown in Table~\ref{table2}, each component improves performance over the baseline across metrics including AUC, FWT, and FAA. RAC significantly improves AUC and FAA, indicating stronger knowledge retention, while the CMI consistently enhances overall performance by preserving cross-modal structure. Combining both components yields further improvements, demonstrating their complementary effects and confirming that Info-VLA achieves a better stability–plasticity trade-off.

\section{Conclusion}

This work investigates catastrophic forgetting in continual VLA learning and identifies the degradation of cross-modal representation structure as a key cause of performance deterioration. As models adapt to new tasks, the alignment among visual observations, language instructions, and actions gradually drifts, leading to unstable policies and inconsistent latent representations.
To address this issue, we propose \textbf{Info-VLA}, an information-preserving continual learning framework that combines replay-anchored contrastive alignment with cross-modal mutual information regularization.
Experiments on the LIBERO benchmark show that Info-VLA consistently outperforms existing continual learning baselines, achieving higher task performance while substantially mitigating catastrophic forgetting. We are currently exploring continual learning experiments on real robotic platforms to further evaluate the practicality of Info-VLA in real-world settings.

\bibliographystyle{splncs04}
\bibliography{main}




\end{document}